\documentclass[10pt]{article}
\usepackage{amsmath,graphicx}
\usepackage{subcaption}
\usepackage[scaled]{helvet}

\usepackage{tabto}
\usepackage[numbib,nottoc]{tocbibind}
\usepackage[a4paper,left=27mm,right=27mm,top=40mm,bottom = 25mm,headsep=22mm,footskip=10mm]{geometry}
\usepackage{booktabs}

\usepackage{fancyhdr}
\usepackage{sectsty}
\sectionfont{\fontsize{14}{10}}
\sectionfont{\MakeUppercase}
\subsectionfont{\fontsize{12}{10}}
\subsubsectionfont{\fontsize{12}{10}}

\makeatletter
\renewcommand\subsubsection{\@startsection{subsubsection}{3}{\z@}%
                                     {-3.25ex\@plus -1ex \@minus -.2ex}%
                                     {1.5ex \@plus .2ex}%
                                     {\normalfont\large\bfseries}}
\makeatother

\makeatletter
\def\@seccntformat#1{\@ifundefined{#1@cntformat}%
   {\csname the#1\endcsname\quad}
   {\csname #1@cntformat\endcsname}}
\newcommand\section@cntformat{\thesection\quad\hspace{0.4047cm}}
\makeatother

\makeatletter
\renewcommand{\maketitle}{\bgroup\setlength{\parindent}{0pt}

\begin{flushleft}
  \fontsize{16pt}{20pt}\selectfont\textbf{\@title} \\
    \vspace{12pt} 
  \fontsize{10pt}{12pt}\selectfont{\@author}
\end{flushleft}\egroup
}
\makeatother
\title{\uppercase{Self-Supervised Learning for Improved Synthetic Aperture Sonar Target Recognition}}
\date{6/2/2023}
\author{%
BW Sheffield \tabto{25mm} Naval Surface Warfare Center, Panama City, USA \\
}

\usepackage{setspace}

\usepackage{cite}

\usepackage{titlesec}
\titlespacing\section{0pt}{10pt}{0pt}
\titlespacing\subsection{0pt}{10pt}{0pt}
\titlespacing\subsubsection{0pt}{10pt}{0pt}
\begin{document}

\maketitle

\section{Introduction}

The application of computer vision in autonomous underwater vehicles (AUVs) presents unique challenges due to the unpredictable and often harsh conditions of marine environments. Traditional computer vision research, which primarily relies on optical camera imagery, struggles to adapt to underwater settings characterized by poor lighting, sediment suspension, and turbidity. Consequently, acoustic sonar, particularly its variant - synthetic aperture sonar (SAS), has emerged as a preferred choice for underwater imaging. SAS-equipped AUVs can sweep seafloors to generate high-resolution imagery, offering a level of detail superior to other sonar types. However, the high-resolution SAS imagery, while rich in detail, is voluminous and presents a significant challenge for labeling, a crucial step for training deep neural networks (DNNs).

DNNs have gained traction in comparison to classical machine learning methods. This is attributed to their ability to autonomously discover features in data. This eliminates the need for manual feature crafting by subject matter experts. A significant limitation of DNNs is their requirement for large volumes of labeled data and considerable computational power to learn robust features. In the context of SAS, not only is labeled data scarce, but it is also not as readily available as it is for conventional camera imagery.

Recently, self-supervised learning (SSL) has gained popularity propelled by the increasing availability of computational power and data. SSL enables models to learn features in data without the need for labels, presenting a potential solution to the data labeling challenge in SAS.

\begin{figure}[ht]
\centering
\includegraphics[width=0.8\textwidth]{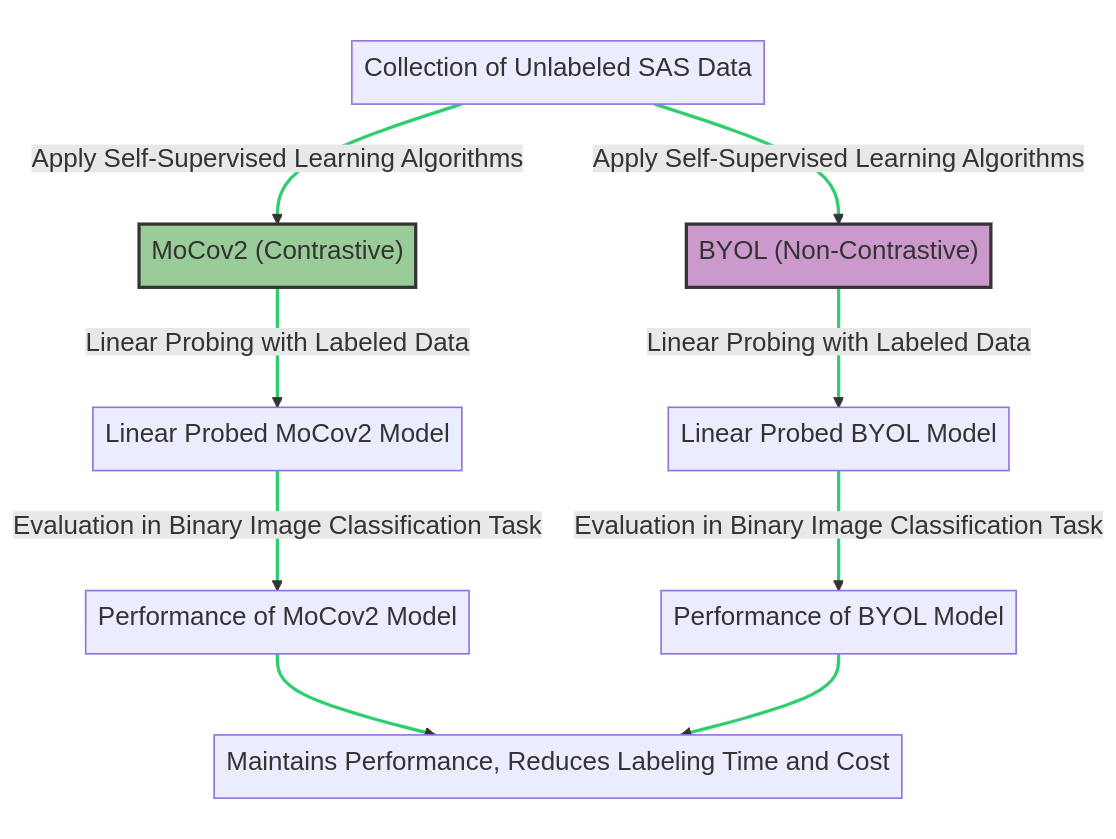}
\caption{The objective of this paper evaluates two different SSL models in limited label scenarios}
\label{fig:conceptual}
\end{figure}

This study aims to evaluate the performance of two prominent SSL algorithms, MoCov2\cite{Chen2020ImprovedBW} and BYOL\cite{grill2020bootstrap}, against the well-regarded supervised learning model, ResNet18\cite{he2016deep}, for the binary image classification task as shown in Figure \ref{fig:conceptual}. The SSL models were trained on real-world SAS data to learn useful feature representations for downstream binary image classification tasks. The findings suggest that while both SSL models can best the performance of a fully supervised model with access to a small amount of labels in limited label scenarios, they do not exceed it when all the labels are used. This study underscores the potential of SSL as a viable alternative to traditional supervised learning, capable of maintaining task performance while reducing the time and costs associated with data labeling.

\section{Related Work}

SSL has been a burgeoning area of research in recent years, particularly within the remote sensing domain \cite{Alosaimi2023SelfsupervisedLF,Berg2022SelfSupervisedLF,Tao2022SelfsupervisedRS,Seneviratne2022UrbanFA,Jung2022ContrastiveSL,Stojnic2021SelfSupervisedLO,Li2022SCLMLNetBF,Wang2022SelfSupervisedVT,Patel2021EvaluatingSA,Wang2022SSL4EOS12AL,Ayush2020GeographyAwareSL}. Although SSL applications have significantly advanced across various fields, their application to SAS remains relatively unexplored.

In 2022, Preciado-Grijalva et al. \cite{preciado2022self} demonstrated the potential of SSL in forward look sonar (FLS) sonar applications can yield classification performance comparable to supervised pre-training in a few-shot transfer learning setup. In the unsupervised and semi-supervised learning domains, researchers have applied methods to reduce the burden of labeled SAS data \cite{chen2017deep,xenaki2022unsupervised,sun2021iterative}. 

The potential of SSL in Synthetic Aperture Radar (SAR) applications, a field closely related to SAS, has been demonstrated in several studies \cite{10.3390/rs14163995,10.1109/mgrs.2022.3198244,10.3390/rs14215500,bourcier2022evaluating,10.3390/rs12111868,10.5194/isprs-annals-v-3-2022-705-2022,10.1109/tgrs.2023.3236664,10.3390/rs12203276,montanaro2022semi}. These studies have shown that SSL can effectively leverage the vast amounts of unlabeled SAR data to achieve meaningful results. 

However, the application of these methodologies remains largely unexplored in the context of SAS data. This gap in the literature may be due to the unique challenges associated with SAS data, such as the sensitive nature of the data and the computational resources required for training.

\section{Methodology}

The subsequent experiments are designed to conduct a comparative evaluation of the performance of the models' representations. This is achieved both qualitatively, through the visualization of the latent spaces, and quantitatively, based on the ultimate classification outcomes. To establish a common baseline, all SSL models, MoCov2 and BYOL, use the same ResNet18 backbone.

Training of the pre-existing models was carried out using PyTorch Lightning for up to 100 epochs. This was carried out on eight Nvidia A6000 GPUs, each equipped with 48GB RAM. The DDP strategy helps to improve the consistency of batch normalization across multiple GPUs. Distributed sampling ensures that each GPU processes a unique subset of the total data in each epoch, leading to more stable training and potentially better performance. Unlike other deep learning methods, high batch sizes are highly desirable in achieving good results as the task of the loss function is to pull positive instances together and push negative instances away.

The downstream task for comparison was binary image classification using binary cross entropy for the loss function. A threshold of 50\% was used to make a decision on whether an image contained a object of interest. In an iterative manner, the SSL pre-trained model is fine-tuned with a percentage of labels with the backbone frozen to compare how well the respective models evaluate against a supervised ResNet18 model. For linear evaluation, early stopping was enabled when training failed to decrease in loss for 10 epochs. 

\subsection{Experimental Setup}

In assessing the efficacy of the representations generated by various SSL frameworks, linear evaluation is used to assess the quality of the learned representations. The linear evaluation method involves training a supervised linear classifier on the SSL pre-trained models, with the model weights kept constant. The classification score derived from this process provides insight into the discriminative capacity of the pre-trained representations and serves as an indirect measure of the model's performance in subsequent tasks \cite{kolesnikov2019revisiting}.

\subsection{Dataset}

Labeled multi-band SAS data is hard to come by and quite limited. Due to the high resolution nature of SAS data, it is often too large for modern GPUs requiring the imagery to be broken up into tiles/snippets/chips. To generate a dataset that consists of snippets, a Reed-Xiaoli\cite{reed1990adaptive} anomaly detector is used to to detect potential objects of interest by extracting snippets from high-resolution SAS imagery. The low and high frequency, call them LF and HF respectively, snippets are first resized to 224x224 each and stacked forming a 2x224x224 multi-band SAS image. Previous works have applied the multi-band approach to success \cite{9976995,emigh2018supervised,Gerg2018MultibandSI}. Different beamformers have been used to generate the SAS imagery providing semantically the same scenes to the human eye yet statistically different. 

The collective snippets make up the four datasets used in experiments: pre-train, train, validation, and test. For simplicity, the labeled datasets(train, validation, and test) used in linear probing experiments are balanced for positive and negative instances.

\subsection{SSL Models and Hyperparameters}

This work leverages two different types of SSL model architectures: MoCov2\cite{Chen2020ImprovedBW} and BYOL\cite{grill2020bootstrap}. The two SSL methods have been categorized as contrastive and non-contrastive.

\begin{itemize}
\item \textbf{MoCov2:} displays strength in learning meaningful representations by contrasting positive and negative samples where distinguishing between different classes is important. However it's contrastive strength, it requires careful selection of negative samples and the size of the queue can significantly affect the performance.

\item \textbf{BYOL:} a popular non-contrastive SSL method as it avoids the need for negative samples which can simplify the training process and reduces computational requirements that contrastive loss functions require such as large batch sizes. The non-contrastive method does come at cost where distinguishing between different classes is crucial.
\end{itemize}

\begin{table}[ht]
\centering
\renewcommand{\arraystretch}{0.4} 
\begin{tabular}{lll}
\toprule
\textbf{Hyperparameter} & \textbf{MoCov2} & \textbf{BYOL} \\ 
\midrule
Backbone                & ResNet18 & ResNet18      \\ 
Channels                & 2 & 2     \\ 
Epochs                  & 100 & 100    \\ 
Optimizer               & AdamW  & AdamW  \\ 
Scheduler               & Cos Anneal & Cos Anneal  \\ 
Loss                    & NTXent  & Neg Cos     \\ 
Learning Rate           & 0.003 & 0.003       \\ 
Weight Decay            & 0.001  & 0.001    \\ 
Batch Size              & 768  & 512         \\ 
\bottomrule
\end{tabular}
\caption{Two different types of SSL models, MoCov2 and BYOL, with similar hyperparameters.}
\label{tab:hyperparameter_table}
\end{table}

\subsection{Data Augmentations}

\begin{figure}[ht]
\centering
\includegraphics[width=0.8\textwidth]{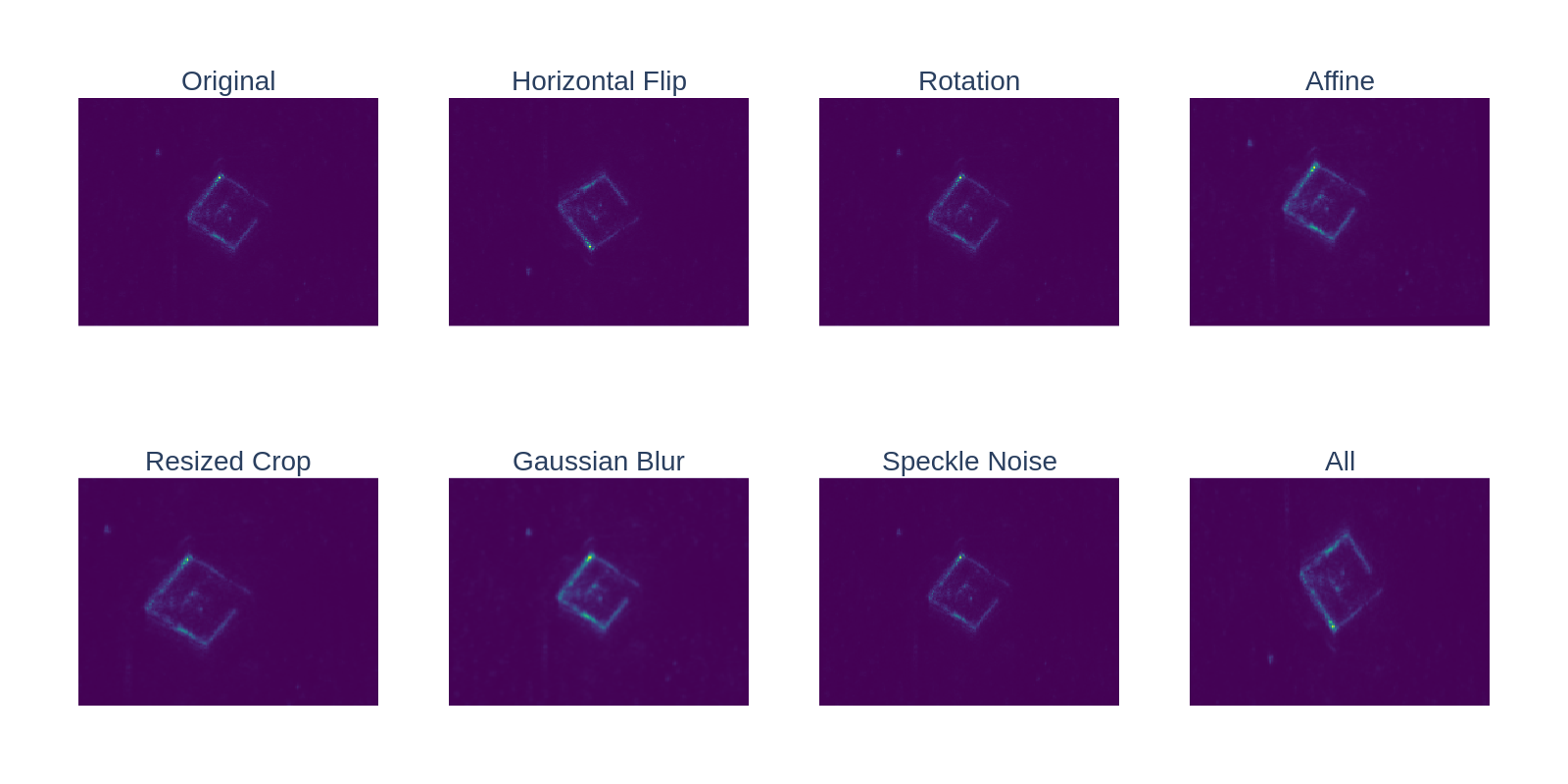}
\caption{Visualization of the SAS data augmentation pipeline for a box-like object during pre-training to create a drastic contrastive image.}
\label{fig:data_augmentations}
\end{figure}

Data augmentation techniques, which generate diverse and challenging examples, are heavily relied upon in SSL. Ensuring that the augmentations are diverse and cover a wide range of transformations can help prevent overfitting thus a moderate amount of augmentations are lightly applied to drive the feature learning process during pre-training as shown in Figure \ref{fig:data_augmentations}. Speckle noise is artificially introduced into the SAS image that multiplies a constant noise factor across the imagery. During training, only horizontal flip augmentations were applied.

\subsection{Performance Metrics}

In the context of image classification with SAS, the evaluation metrics and benchmarks used for binary image classification tasks need to effectively measure the ability of the model to accurately distinguish between objects and objects not of interest (typically representing seafloor clutter or other underwater objects). The following performance metrics were used to evaluate the models:

\begin{itemize}
\item \textbf{Contrastive loss:} During pre-training, the contrastive loss is tracked on a validation dataset providing insight on how well the model is learning to distinguish between similar and dissimilar samples.

\item \textbf{Recall (Sensitivity or True Positive Rate):} This measures the proportion of actual positives (objects of interest) that were identified correctly. A high recall is crucial in image classification because failing to identify an object (false negative), could be disastrous in contested military waters.

\item \textbf{Precision (Positive Predictive Value):} This measures the proportion of positive identifications (identified objects) that were actually correct. A high precision means a low false positive rate, which is desirable in image classification tasks to avoid wasting resources on false detections.

\item \textbf{Area Under the Receiver Operating Characteristic (ROC) Curve (AUC-ROC):} This metric provides a comprehensive measure of performance across all possible classification thresholds, summarizing the trade-off between the true positive rate and false positive rate.

\item \textbf{Accuracy:} This is the simplest metric, representing the proportion of total predictions that were correct. However, accuracy can be misleading if the classes are imbalanced (e.g., if objects are much less common than distractor objects).
\end{itemize}

\section{Results}

\begin{figure}[ht]
\centering
\includegraphics[width=0.6\textwidth]{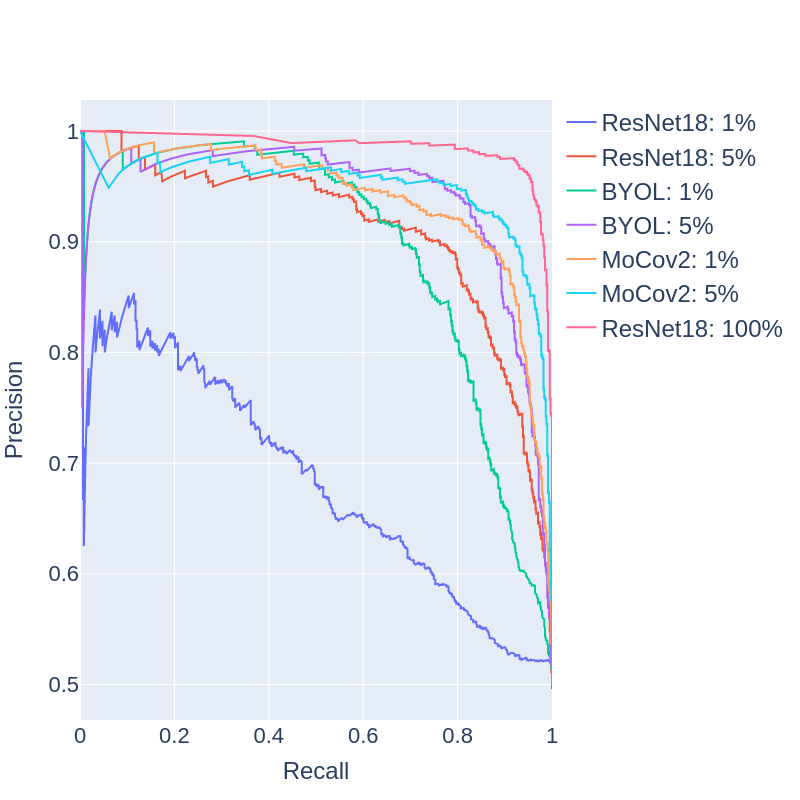}
\caption{Precision-Recall curves demonstrate SSL model trade-offs for varying labeled scenarios.}
\label{fig:pr_curves}
\end{figure}

\begin{figure}[ht]
\centering
\includegraphics[width=0.6\textwidth]{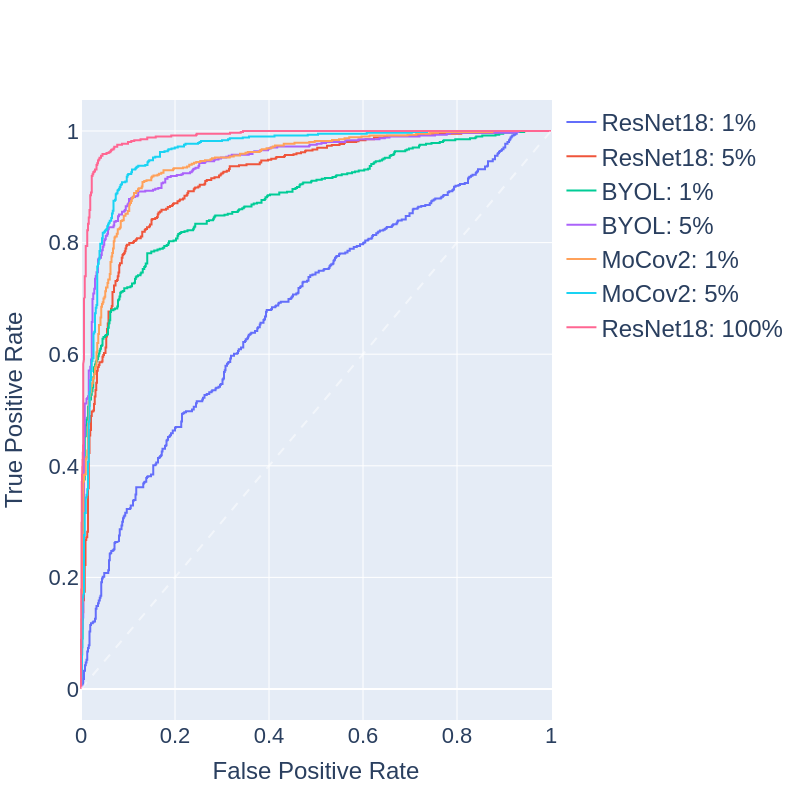}
\caption{ROC curves show SSL model trade-offs for varying labeled scenarios.}
\label{fig:roc_curves}
\end{figure}


\begin{figure}[ht]
\centering
\includegraphics[width=0.6\textwidth]{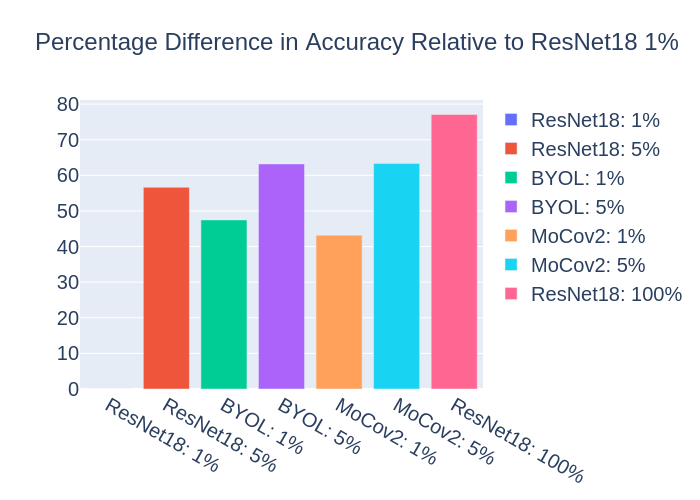}
\caption{Relative accuracy to ResNet18 1\% increases for all other models during evaluation.}
\label{fig:relative_test_acc}
\end{figure}

The integration of SSL into SAS significantly enhances the performance of the SSL models, specifically MoCov2 and BYOL, when only 1\% and 5\% of the labels are utilized during training. However, when compared to the ResNet18 model, which had access to 100\% of the labels, the SSL models fell short, as shown in Figures \ref{fig:pr_curves}, \ref{fig:roc_curves}, and \ref{fig:relative_test_acc}. The SSL algorithms were able to effectively extract high-level features from the SAS data, resulting in enhanced performance in downstream tasks for limited label scenarios.

\section{Discussion}

The results suggest that SSL can be effectively applied to SAS, similar to its successful application in SAR and other computer vision tasks. The improved performance in image classification tasks indicates the potential of SSL in enhancing SAS target recognition for low labeled regimes. However, when abundant data labels exist, supervised learning outperforms in all aspects.

\begin{figure}[ht]
\centering
\includegraphics[width=0.6\textwidth]{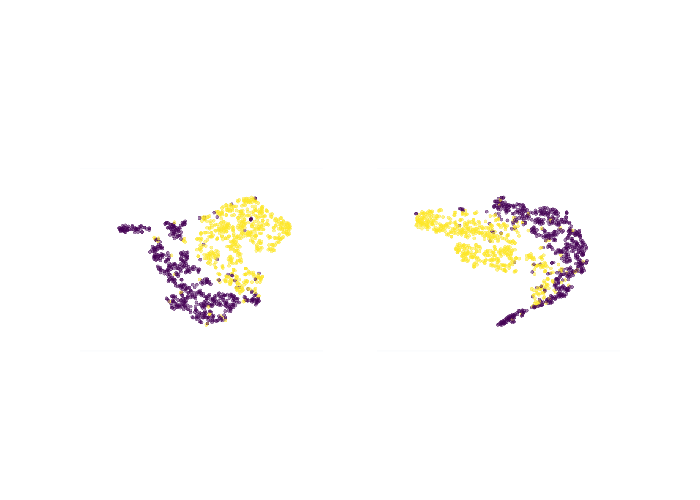}
\caption{MoCov2 and BYOL t-SNE representations show how well each model clusters SAS images.}
\label{fig:ssl-tsne}
\end{figure}

In order to better understand the feature representations learned by the models, t-SNE \cite{van2008visualizing}, a popular technique for visualizing high-dimensional data was deployed. Figure \ref{fig:ssl-tsne} shows the t-SNE visualizations of the feature representations learned by MoCov2 and BYOL.

As can be seen from the visualizations, both the SSL models and the supervised model have learned to cluster the sonar images in a meaningful way, with images of the same class clustering together. This suggests that the models have learned to extract features that are relevant for the task of sonar object classification. More compact and well-separated clusters indicate the SSL models learned robust and discriminative features, which could potentially lead to better performance in downstream tasks other than image classification such as object detection, segmentation, and change detection.

\subsection{Implication of Results}


Based on the findings, the application of SSL to SAS significantly improves the performance of target recognition tasks for low labeled regimes. This has several important implications.

Firstly, it suggests that SSL can effectively leverage the abundance of unlabeled SAS data, which has traditionally been a challenge in this field. This could potentially revolutionize the way we process and analyze SAS data, leading to more efficient and cost-effective methods.

Secondly, the improved performance in downstream tasks such as image classification indicates that SSL can enhance the practical utility of SAS in various applications, such as underwater exploration, marine archaeology, and other naval applications.

Finally, the results contribute to the growing body of evidence supporting the use of SSL in remote sensing and could stimulate further research in this area.

\section{Conclusion}

The potential of self-supervised learning to improve the classification of SAS images is underscored in this study. Given their success in various computer vision tasks, future research could explore the use of Vision Transformers (ViTs) as backbones for SSL with SAS data. 

Additionally, a multi-modal SSL approach that leverages all available data collected by autonomous underwater vehicles, such as bathymetric data or other sonar modalities, could potentially provide richer representations and improve performance.  

While the application of SSL to SAS tasks is promising, it is still in its infancy. Further exploration could significantly advance automated underwater computer vision tasks.

\bibliographystyle{unsrt} 
\bibliography{sources}

\end{document}